\documentclass[letterpaper,journal]{IEEEtran}
\usepackage{amsmath,amsfonts}
\usepackage{algorithmic}
\usepackage{algorithm}
\usepackage{array}
\usepackage[caption=false,font=normalsize,labelfont=sf,textfont=sf]{subfig}
\usepackage{textcomp}
\usepackage{stfloats}
\usepackage{url}
\usepackage{verbatim}
\usepackage{graphicx}
\usepackage{hyperref}
\usepackage{booktabs}
\usepackage{tabularx}

\begin{document}

\title{Listen Before Act: Audio-Conditioned Robot Manipulation with ListenVLA}

\author{{Jiaqi Fan, Rui Xu, Jiakun Xia, Xuyu Fan, Shuailei Ma, Yuefeng Wang, Qing Wang, Jiarui Zhang, Yili Fu, Xu Li}
\thanks{}
\thanks{}}



\maketitle

\begin{abstract}
Vision–language–action (VLA) models have advanced robotic manipulation by connecting visual perception and language instructions to action generation. However, many manipulation tasks depend on auditory cues that cannot be reliably inferred from visual observations and may need to be elicited through physical interaction. To address this gap, we present ListenVLA, an audio-conditioned VLA policy built on GR00T N1.7 for sound-guided manipulation. ListenVLA combines semantic audio embeddings with waveform-derived acoustic cues over a causal audio history and conditions continuous action generation through audio tokens and feature modulation. Moreover, we introduce an audio–action binding objective that uses context-matched demonstrations to link auditory differences to demonstrated action differences. The objective supervises short-horizon differences between flow-velocity predictions under different audio conditions, with shared pure-noise action inputs and fixed non-audio observations, alongside full-chunk action supervision. Furthermore, we develop LIBERO-Audio, an eight-task benchmark covering sound-based target selection, responses to auditory events, and physical interactions that reveal acoustic evidence. Experiments on LIBERO-Audio and three real-world manipulation tasks show excellent performance. Ablation studies support the contributions of binding supervision and audio modulation, while audio swap tests show that sound influences action predictions. Our code and videos are available at \url{https://anonymous.4open.science/r/listenvla-F18D/}.
\end{abstract}


\section{Introduction}
\begin{figure}[t]
    \centering
    \includegraphics[width=\columnwidth]{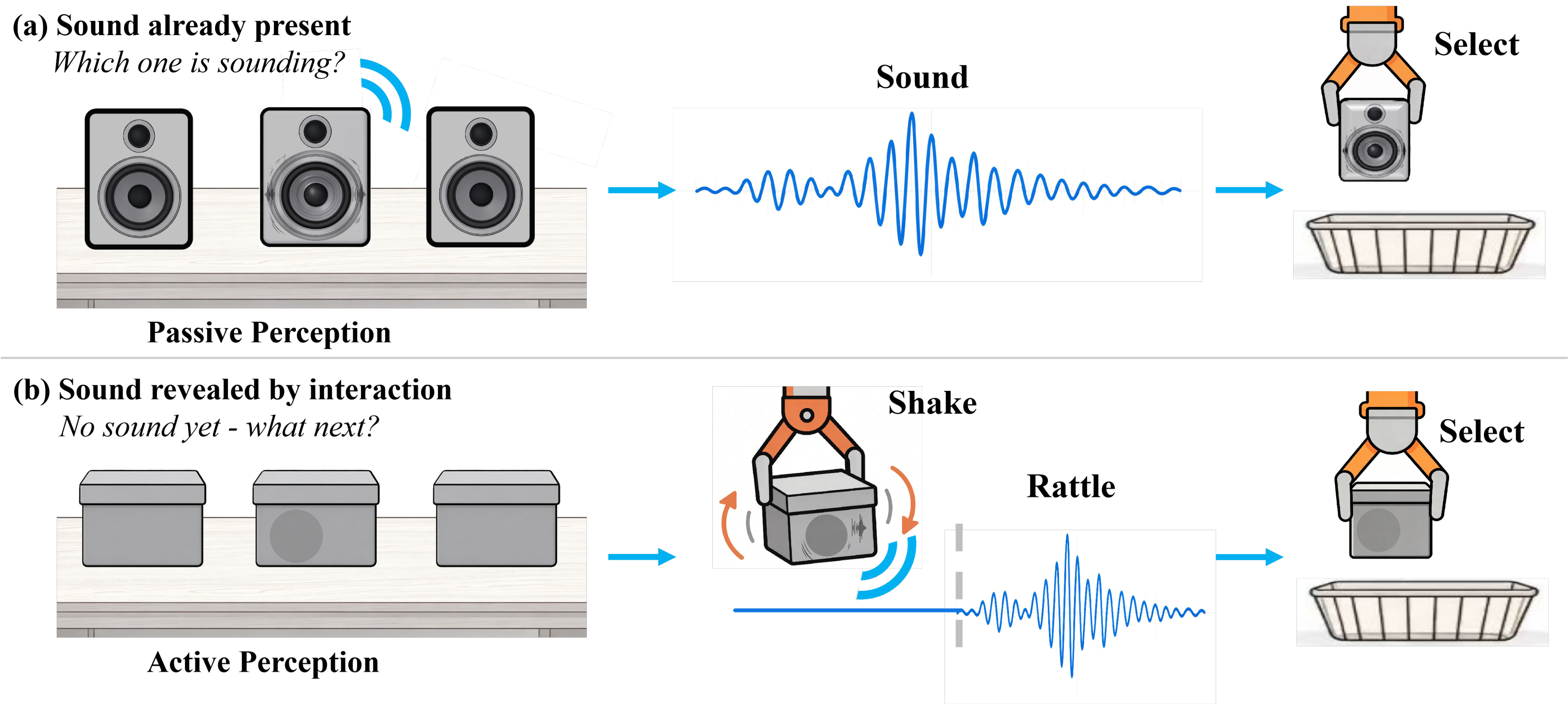}
    \caption{Representative sound-guided manipulation tasks in LIBERO-Audio. (a) When sound is already present, the robot listens to distinguish the sounding object from visually identical alternatives and selects it. (b) When objects are initially silent, the robot must interact with them to obtain auditory evidence: grasping and shaking a box reveals a rattle that identifies the target.}
    \label{fig:task_overview}
\end{figure}

\IEEEPARstart{E}{veryday} interaction with the physical world is guided by more than what we can see. Sound often provides cues for directing attention, timing actions, or probing hidden states. Sometimes this evidence arrives on its own; at other times, we must create it through interaction. Fig.~\ref{fig:task_overview} illustrates this distinction: a robot can listen to select a sounding object, whereas identifying a rattling box requires first grasping and shaking it. Bringing these abilities to robots requires connecting auditory perception with the decisions made during manipulation. A robot must relate a sound to the objects around it, interpret its relevance to the instruction, and use that evidence to guide its next action. This raises a central question: how can a robot policy use auditory evidence to resolve action ambiguity when vision alone is insufficient?

Vision--language--action (VLA) models provide a promising starting point for this problem. OpenVLA transfers pretrained visual and linguistic representations to robot action prediction, while $\pi_0$ and GR00T combine vision--language backbones with continuous action generation through flow matching~\cite{openvla,pi0,groot}. These approaches connect language-conditioned perception to control, but their standard formulations do not use environmental audio as an observation. Complementary sensing has therefore become an important direction. ManiWAV learns manipulation policies from synchronized audio--visual demonstrations, using auditory feedback when visual observations are ambiguous or incomplete~\cite{ManiWAV}. FuSe extends this direction to pretrained generalist policies, incorporating additional sensory inputs, including sound, through language grounding~\cite{Fuse}. Audio-VLA integrates contact audio into a VLA model to capture interaction feedback during manipulation~\cite{Audio-vla}. HEAR maintains causal audio context and uses future audio prediction during training to support manipulation guided by sounds that evolve over time~\cite{hear}. Together, these studies highlight the potential of auditory perception to complement vision and inform robot actions.

Despite this progress, learning to translate acoustic observations into appropriate manipulation behavior remains challenging. In sound-guided tasks, the relevance of an acoustic cue depends on both the instruction and the robot's ongoing interaction. A sound may identify which object to retrieve, indicate when to begin moving, or reveal whether an exploratory action has uncovered the desired object. The policy must therefore relate what it hears to the current stage of the task and adjust its behavior accordingly. Learning this relationship from demonstrations can be difficult: trajectories associated with different acoustic conditions often share similar reaching and probing motions, yet require different actions at the moment of decision. When training is dominated by these shared movements, a policy may learn the overall motion pattern without reliably learning when sound should change it. The challenge is therefore to adapt a pretrained VLA so that auditory evidence reliably guides action selection.

To address this challenge, we propose ListenVLA, an audio-conditioned manipulation policy built on GR00T N1.7. ListenVLA combines semantic audio features with measurements of sound energy and its changes over recent observations, and uses the resulting representation to condition action generation. We also introduce an audio–action binding objective to teach the policy how its actions should change with sound. We select demonstration pairs with similar visual and robot states but different sounds and subsequent actions. For each pair, we hold the non-audio inputs fixed and make two predictions from the same initial noise using different audio inputs. Their difference is supervised by the demonstrated action difference over a short horizon, while standard action supervision is retained over the full action chunk.

Moreover, we introduce LIBERO-Audio, an eight-task benchmark built on LIBERO~\cite{libero}, to evaluate how robots use sound during manipulation. The tasks cover selecting objects by sound, following instructions that distinguish sound categories, waiting for auditory events, and probing objects to obtain acoustic evidence. The target-selection tasks use visually identical candidate objects, making sound necessary to distinguish their relevance to the instruction. We complement closed-loop task evaluation with controlled audio-swap tests, which replace the audio input while keeping the images, instruction, current robot state, and initial action noise fixed. These tests measure whether sound affects the predicted actions, while closed-loop evaluation measures task completion.

On LIBERO-Audio, ListenVLA substantially outperforms the compared models, while ablation studies support the effectiveness of the complete method. We further validate ListenVLA on three real-world manipulation tasks, showing that the approach also supports manipulation with physical auditory feedback.

Our contributions are threefold:
\begin{itemize}
    \item We propose ListenVLA, an audio-conditioned manipulation policy based on GR00T N1.7. It combines semantic audio features with temporal acoustic measurements to condition continuous action generation.
    \item We introduce an audio–action binding objective that uses demonstrations with matched visual and robot context to supervise how predicted actions change with audio. By fixing the non-audio observations and starting both predictions from the same pure noise, the objective links auditory differences to demonstrated action differences over a short horizon.
    \item We introduce LIBERO-Audio, an eight-task benchmark for sound-guided manipulation. Comparisons with existing audio-based policies, ablation studies, controlled audio-swap tests, and experiments on three real-world manipulation tasks evaluate the proposed method.
\end{itemize}

\section{Related Work}

\subsection{Vision--Language--Action Models}
Vision--language--action models build on language-conditioned control and embodied multimodal learning~\cite{rt1,palme,rt2}. Octo and OpenVLA learn generalist policies from diverse robot demonstrations~\cite{octo,openvla}. Action generation uses flow matching in $\pi_0$ and GR00T, diffusion decoding in TinyVLA, or parallel chunk prediction in OpenVLA-OFT~\cite{pi0,groot,tinyvla,openvla_oft}. Language-based action hierarchies and heterogeneous co-training support execution and generalization~\cite{rth,pi05}. These methods primarily condition on images and language. ListenVLA extends GR00T N1.7 with auditory conditioning for manipulation tasks in which visual observations alone do not identify the required action.

\subsection{Auditory Perception for Robotic Manipulation}
Sound reveals object properties and interaction events that may be difficult to infer visually. SonicSense infers object properties and container contents from exploratory acoustic vibrations~\cite{sonicsense}. See, Hear, and Feel fuses vision, contact audio, and touch for dense packing and pouring, while ManiWAV learns from synchronized audio--visual demonstrations collected with an ear-in-hand device~\cite{see_hear_feel,ManiWAV}. FuSe adapts generalist policies to additional sensory inputs through language grounding~\cite{Fuse}, and Audio-VLA incorporates contact audio into a VLA~\cite{Audio-vla}. Audio world models predict future sounds to guide manipulation~\cite{audio_world_models}. HEAR combines streaming auditory memory and future-audio prediction with flow-matching action generation~\cite{hear}. ListenVLA combines temporal audio conditioning with an audio--action binding objective. Context-matched demonstrations supervise short-horizon prediction differences under audio substitution, while non-audio observations and the initial noise are held fixed.

\subsection{Benchmarks for Sound-Guided Manipulation}
RLBench and LIBERO provide manipulation task suites and evaluations of knowledge transfer~\cite{rlbench,libero}. ObjectFolder Benchmark studies object recognition, reconstruction, and manipulation with vision, audio, and touch~\cite{objectfolder_benchmark}. Audio-VLA adds collision-triggered audio to RLBench and LIBERO~\cite{Audio-vla}, while HEAR-Bench evaluates manipulation under acoustic timing constraints~\cite{hear}. LIBERO-Audio comprises eight tasks covering sound-based target selection, instruction-dependent sound recognition, delayed auditory events, and interactions that reveal or suppress acoustic cues. Its target-selection tasks use visually identical candidates, requiring auditory evidence to distinguish the target.

\section{Method}
\subsection{Problem Formulation and Overview}
We study language-conditioned manipulation using images, proprioception, and sound. At time \(t\), the observation comprises external and wrist-view images \(I_t\), current robot state \(s_t\), instruction \(\ell\), and waveform history \(W_t\). Audio histories end at the observation timestamp. The policy predicts a continuous action chunk \(\hat A_t\sim\pi_\theta(\cdot\mid I_t,s_t,\ell,W_t)\), supervised by demonstrations \(A_t=(a_t,\ldots,a_{t+H-1})\).

\begin{figure*}[!t]
\centering
\includegraphics[width=\textwidth]{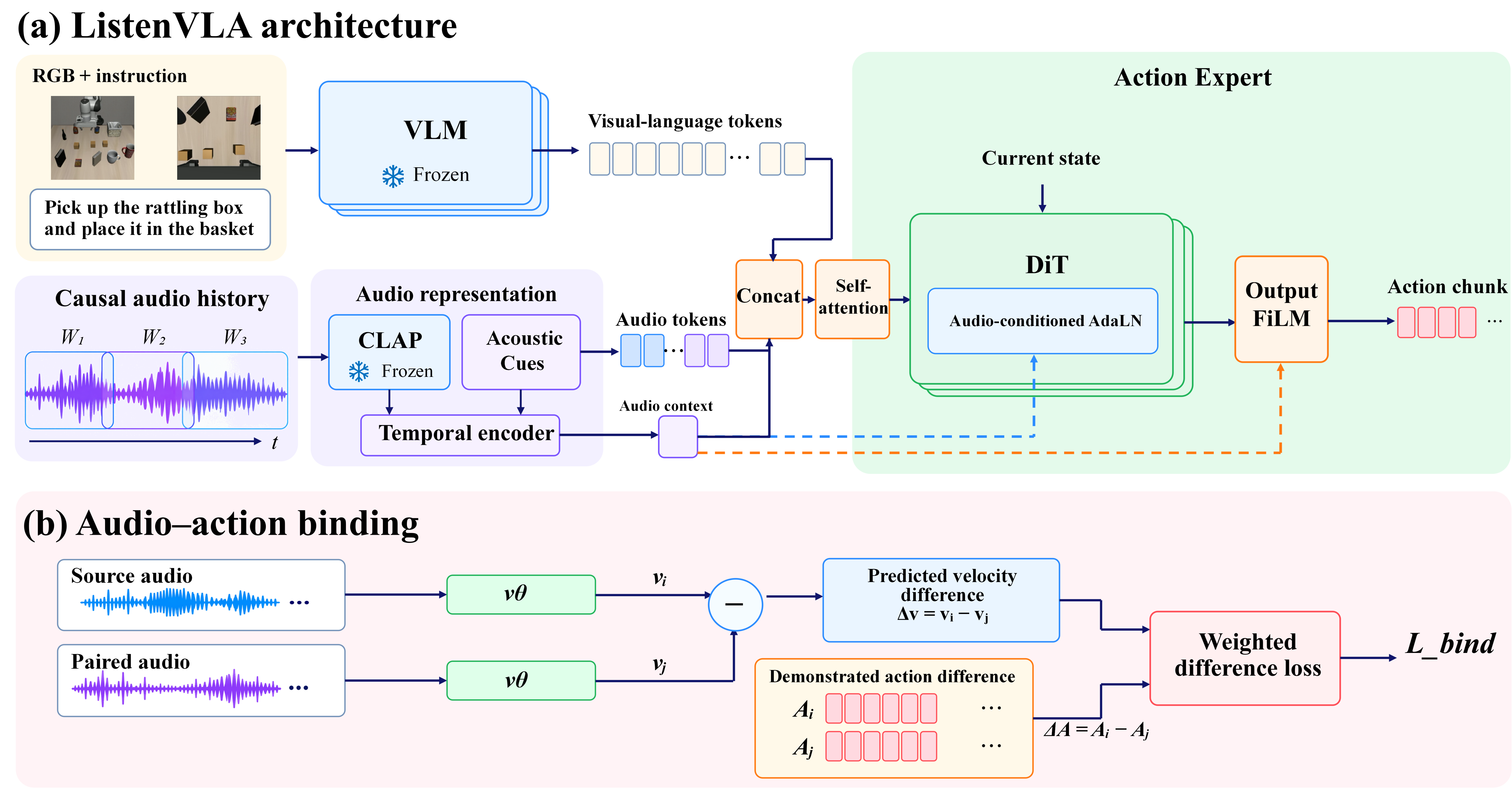}
\caption{ListenVLA architecture and audio--action binding. (a) Three overlapping windows form the causal audio history. Frozen CLAP embeddings and waveform-derived acoustic cues supply window tokens and feed a temporal encoder. The resulting audio context supplies additional tokens and conditions AdaLN within the DiT (blue dashed path) and output FiLM before action decoding (orange dashed path). Audio tokens are concatenated with frozen VLM tokens and fused through self-attention. (b) Training-only binding uses two predictions from the same policy, with identical non-audio inputs.}
\label{fig:listenvla}
\end{figure*}

ListenVLA extends GR00T N1.7 with temporal audio conditioning and an audio--action binding objective. Fig.~\ref{fig:listenvla}(a) shows how audio tokens, audio-conditioned adaptive layer normalization (AdaLN), and output FiLM condition the action expert. During training, context-matched demonstrations supervise the response to controlled audio substitution (Fig.~\ref{fig:listenvla}(b)). We also introduce LIBERO-Audio, an eight-task benchmark for sound-guided manipulation. The tasks cover sound-based target selection, responses to auditory events, and physical interactions that reveal acoustic evidence.

\subsection{ListenVLA Architecture}
\subsubsection{Temporal audio representation}
Sound category helps identify a target, while changes in intensity provide feedback during approach and interaction. We combine CLAP semantic embeddings~\cite{clap} with waveform-derived acoustic cues measuring energy and its temporal changes. The past two seconds are divided into three overlapping one-second windows, shown as \(W_1,W_2,W_3\) in Fig.~\ref{fig:listenvla}(a). For each window \(w_k\), CLAP produces \(c_k\), and the acoustic branch computes
\begin{equation}
\begin{aligned}
r_k&=[R_k,B_k,\Delta R_k,\Delta B_k],\\
u_k&=P_f([P_c(c_k);P_r(\bar r_k)])+e_k.
\end{aligned}
\label{eq:audio-representation}
\end{equation}
Here, \(R_k\) is log RMS amplitude and \(B_k\) is log energy in the 200–4000 Hz band; differences are computed between adjacent valid windows. Training-set statistics standardize these acoustic cues, followed by clipping to \([-5,5]\), yielding \(\bar r_k\). Learned projections fuse the two representations, and \(e_k\) encodes relative window age. The temporal encoder uses a two-layer transformer followed by masked mean pooling and projection to produce audio context \(z_t\). Incomplete windows are masked, including differences that require them.
\subsubsection{Conditioning the action network}
As shown in Fig.~\ref{fig:listenvla}(a), GR00T's vision--language backbone produces \(V_t=F_{\mathrm{VL}}(I_t,\ell)\). We project each window's CLAP embedding and acoustic cues into separate tokens, and \(z_t\) into twelve context tokens. These window-token projections are independent of the projections inside the temporal encoder. The resulting eighteen audio tokens are concatenated with \(V_t\) along the sequence dimension. Feature normalization and self-attention fuse this sequence after the backbone, yielding context \(C_t\) for the DiT action transformer within the action expert. Audio tokens enter its non-image attention branch. Window tokens retain their validity masks, and context tokens are active when any window is valid.

Audio context \(z_t\) also conditions two distinct modulation paths. A residual added to the flow-time embedding conditions AdaLN within the DiT, while output FiLM~\cite{film} modulates its final features before action decoding:
\begin{equation}
\begin{aligned}
e'_\tau&=e_\tau+\alpha\tanh(P_\tau(z_t)),\\
h'&=(1+\gamma(z_t))\odot h+\beta(z_t).
\end{aligned}
\label{eq:audio-modulation}
\end{equation}
The embedding \(e'_\tau\) supplies the AdaLN condition (blue dashed path in Fig.~\ref{fig:listenvla}(a)). Output FiLM applies the audio-conditioned scale and shift to features \(h\), yielding \(h'\) for the action decoder (orange dashed path). We use \(\gamma=\eta\tanh(P_\gamma(z_t))\), \(\beta=\eta P_\beta(z_t)\), \(\alpha=0.1\), and \(\eta=0.5\). Both modulation paths are gated by audio validity. Here, \(\tau\) denotes flow time and \(t\) denotes observation time.

We freeze the pretrained vision–language backbone and CLAP, and train the audio modules, action-side fusion, and action network. The resulting conditional velocity field uses the current robot state together with the token and modulation paths.

\subsection{Audio–Action Binding Training}
We train on demonstration dataset \(\mathcal D\) with full-chunk flow-matching supervision~\cite{flow_matching} and an auxiliary audio--action binding objective. Let \(o=(I,s,\ell)\) collect non-audio inputs, and let \(v_\theta(X,\tau;o,W)\) denote the complete conditioned velocity field. For a normalized demonstration chunk \(A\) and Gaussian noise \(\epsilon\), we form \(X_\tau=(1-\tau)\epsilon+\tau A\) and minimize
\begin{equation}
\mathcal L_{\mathrm{all}}=
\mathbb E_{\mathcal D,\epsilon,\tau}
\left[\left\langle
[v_\theta(X_\tau,\tau;o,W)-(A-\epsilon)]^2
\right\rangle_M\right].
\label{eq:action-supervision}
\end{equation}
The brackets average over valid action elements specified by mask \(M\); the expectation averages over examples, noise, and flow times. The audio--action binding objective supervises audio-induced differences in flow-velocity predictions using demonstrated action differences from context-matched pairs.

\subsubsection{Context-matched pairs}
We mark candidate anchors offline near motion starts, acoustic transitions, and translational action reversals. Acoustic-event windows start at the event; motion-event neighborhoods may include preceding frames. For each source anchor \(i\), we search a bounded set of other demonstrations of the same task. Candidate donors must match the source within 12 mm in
end-effector position, 0.20 rad in SO(3) orientation, and
4 mm in gripper opening. We additionally compare relative
end-effector trajectories over the preceding four steps,
requiring a root-mean-square difference below 12 mm and
matching moving/stationary and gripper-phase indicators.
For each camera, RGB images are normalized to $[0,1]$ and
average-pooled to $8\times8$ fingerprints; pairs are retained only when their root-mean-square difference does not exceed 0.20 in either view. Motion history is used only for pair selection; the policy receives the current robot state.Motion history is used for matching, while the policy receives current state.

Candidates must meet a minimum mean absolute difference in their first-eight-step normalized actions and a minimum separation in acoustic cues or coarse spectral content. Among accepted candidates, we choose the donor minimizing context distance \(D_{ij}\): the mean tolerance-normalized position, rotation, opening, and motion distances, plus mean visual fingerprint distance. Confidence is \(c_{ij}=\exp(-D_{ij})\). 

\subsubsection{Controlled audio substitution}
For each context-matched pair, we evaluate the policy twice at the pure-noise endpoint \(\tau=0\), using a shared Gaussian sample \(\tilde\epsilon_i\). Both predictions use the same policy parameters, source images, instruction, and current robot state, differing only in their audio histories and associated features and masks. As shown in Fig.~\ref{fig:listenvla}(b), we supervise the resulting flow-velocity difference with the demonstrated action difference over the first eight steps:
\begin{equation}
\begin{aligned}
\Delta v_{ij}={}&v_\theta(\tilde\epsilon_i,0;o_i,W_i)\\
&-v_\theta(\tilde\epsilon_i,0;o_i,W_j),\\
\ell_{ij}^{\mathrm{bind}}={}&
\left\langle[\Delta v_{ij}-(A_i-A_j)]^2
\right\rangle_{M_{ij}^{(8)}}.
\end{aligned}
\label{eq:binding}
\end{equation}
Here, \(\Delta v_{ij}\) is the difference between the policy's flow-velocity predictions under the source and paired audio conditions. The target \(A_i-A_j\) is the difference between their normalized demonstration action chunks. At \(\tau=0\), each target velocity has the form \(A-\tilde\epsilon_i\), so the shared noise cancels upon subtraction. The mask \(M_{ij}^{(8)}\) selects action elements valid in both demonstrations within the first eight steps. The operator \(\langle\cdot\rangle_{M_{ij}^{(8)}}\) averages the squared error over these elements.

Both auxiliary forwards disable ordinary and state dropout, with gradients propagated through both branches.

\subsubsection{Pair weighting and total objective}
We weight supervision by context confidence and acoustic separation:
\begin{equation}
\begin{aligned}
w_{ij}=c_{ij}\max\big\{&
\operatorname{clip}_{[0,1]}(d_{ij}^{\mathrm{ac}}/0.25),\\
&\operatorname{clip}_{[0,1]}(d_{ij}^{\mathrm{CLAP}}/0.05)\big\}.
\end{aligned}
\label{eq:pair-weight}
\end{equation}
Over aligned, jointly valid windows, \(d_{ij}^{\mathrm{ac}}\) is the RMS difference between standardized acoustic cues before clipping, and \(d_{ij}^{\mathrm{CLAP}}\) is mean cosine distance between frozen CLAP embeddings. Weights are detached from differentiation. For accepted pairs \(\mathcal P\) in a batch with \(N_{\mathrm{anchor}}\) candidate anchors, training minimizes
\begin{equation}
\mathcal L=
2\mathcal L_{\mathrm{all}}+
\underbrace{\frac{\sum_{(i,j)\in\mathcal P}
w_{ij}\ell_{ij}^{\mathrm{bind}}}
{\max(1,N_{\mathrm{anchor}})}}_{\mathcal L_{\mathrm{bind}}}.
\label{eq:total-objective}
\end{equation}
The denominator includes anchors without accepted donors; their binding contribution is zero. Coefficients remain fixed throughout training. Pair selection and auxiliary predictions are training-only; deployment uses a single conditioned policy.

\subsection{LIBERO-Audio Benchmark}
\label{sec:libero-audio}

LIBERO-Audio extends LIBERO~\cite{libero} with eight tasks covering sound-based selection, instruction-dependent sound recognition, delayed response, and interaction to reveal acoustic evidence. Seven tasks contain three visually identical candidates within a category; T6 uses a single clock. Target locations vary across episodes, and every task requires placing the designated object in a basket.
\begin{table*}[!t]
\centering
\caption{Task definitions in LIBERO-Audio.}
\label{tab:tasks}
\begin{tabularx}{\textwidth}{l l X}
\toprule
Task & Acoustic setting & Required behavior \\
\midrule
T1: Sounding speaker      & One beeps; two are silent                              & Select the sounding speaker \\
T2: Beeping speaker       & Beep, music, and barking play simultaneously           & Select beep \\
T3: Music speaker         & Same sound mixture as T2                               & Select music \\
T4: Barking speaker       & Same sound mixture as T2                               & Select barking \\
T5: Rattling box          & Initially silent; the target rattles when grasped and shaken & Trigger the target's grasp-and-shake event \\
T6: Delayed alarm         & One clock rings after a delay                          & Wait for the alarm before engaging \\
T7: Sounding buzzer       & One buzzer initially sounds                            & Silence the target \\
T8: Probe-and-silence buzzer & Pressing the initially silent target activates a warning & Activate and then silence the target \\
\bottomrule
\end{tabularx}
\end{table*}

T2–T4 require selecting the instructed sound category from the same multi-source acoustic scene. T5 and T8 reveal target information through interaction.

The environment uses a Franka Panda with external and wrist cameras and a virtual monaural microphone at the end effector. Audio mixes source clips with distance attenuation, microphone directivity, propagation delay, and a simplified tabletop reflection. Simulator events control playback, including grasp-and-shake triggers, delayed alarms, and button presses.

We collect synchronized listening and probing demonstrations at 20 Hz, with audio stored at 16 kHz. Policy observations contain only images, robot state, instructions, and audio. Success combines target placement with the events in Table~\ref{tab:tasks}.

\section{Experiments}
We evaluated ListenVLA on LIBERO-Audio and three real-world manipulation tasks. We compared task completion with existing audio-conditioned policies and examined the contributions of the binding objective and audio conditioning paths. An audio swap test assessed the influence of sound beyond chance target selection, with fixed non-audio observations and sampling noise.

\subsection{Experimental Setup}

\subsubsection{Simulation and training}
LIBERO-Audio contains the eight tasks described in Sec.~\ref{sec:libero-audio}. The demonstration set comprised 960 episodes: 60 for each of T1--T4 and 180 for each of T5--T8. We fine-tuned ListenVLA from GR00T N1.7 for 100{,}000 updates with a batch size of 16. Training used AdamW with a learning rate of $5\times10^{-5}$, cosine decay, a 5\% warm-up period, and weight decay of $10^{-5}$. The vision--language backbone and CLAP remained frozen. Training used the objective in Eq.~\eqref{eq:total-objective}.

\subsubsection{Baselines}
We adapted HEAR~\cite{hear} and ManiWAV~\cite{ManiWAV} to LIBERO-Audio using demonstrations from the same task suite. HEAR retained its audio-history processing and future-audio prediction objective. We trained a separate ManiWAV policy for each task, following its formulation without language input. ManiWAV used the external camera, while HEAR and ListenVLA used external and wrist views. HEAR and ManiWAV were trained under the same training
budget as ListenVLA.All methods were evaluated on the same set of task initializations.

\subsubsection{Evaluation}
For the baseline comparison reported in Table~\ref{tab:libero-audio}, we evaluated each method over 50 closed-loop episodes per task. ListenVLA and HEAR executed 16 actions per policy query at 20~Hz, with a limit of 2{,}000 control steps. Success followed the task-specific criteria in Sec.~\ref{sec:libero-audio}. 

\subsection{Comparison on LIBERO-Audio}
ListenVLA achieved a mean success rate of 66.25\%, compared with 33.00\% for HEAR and 13.25\% for ManiWAV (Table~\ref{tab:libero-audio}). It obtained the highest success rate on seven tasks. T5 and T8 require interaction to obtain auditory evidence: the robot must grasp and shake a box in T5 and press a button to activate a warning sound in T8. These tasks remained more difficult, with success rates of 32.0\% and 20.0\%, respectively.

\begin{table}[t]
\centering
\caption{Closed-loop success rates on LIBERO-Audio (\%). Each task contains 50 evaluation episodes. Mean is the unweighted average over T1--T8; bold indicates the highest value in each row.}
\label{tab:libero-audio}
\resizebox{0.7\linewidth}{!}{
\begin{tabular}{lccc}
\toprule
Task & ManiWAV & HEAR & ListenVLA \\
\midrule
T1 & 12.0 & 24.0 & \textbf{64.0} \\
T2 & 34.0 & 20.0 & \textbf{96.0} \\
T3 & 22.0 & 40.0 & \textbf{86.0} \\
T4 & 36.0 & 26.0 & \textbf{94.0} \\
T5 & 2.0 & 18.0 & \textbf{32.0} \\
T6 & 0.0 & \textbf{96.0} & 94.0 \\
T7 & 0.0 & 38.0 & \textbf{44.0} \\
T8 & 0.0 & 2.0 & \textbf{20.0} \\
\midrule
Mean & 13.25 & 33.00 & \textbf{66.25} \\
\bottomrule
\end{tabular}
}
\end{table}

\subsection{Ablation Study}
Table~\ref{tab:ablation} compares four configurations. A is GR00T without audio, B removes audio--action binding, C removes audio modulation, and D is the complete ListenVLA model.

B retains the complete audio-conditioning architecture and is trained solely with full-chunk flow-matching action supervision. The action-loss coefficient remains unchanged, isolating the contribution of the auxiliary binding objective. C retains audio tokens, temporal encoding, and binding, but jointly removes audio-conditioned AdaLN modulation and output FiLM.

Removing binding reduced mean success from 66.25\% to 38.25\%. Jointly removing audio-conditioned AdaLN and output FiLM reduced it to 59.25\%. These results support the contributions of binding supervision and audio modulation to the complete policy. 

\begin{table}[t]
\centering
\caption{Ablation study on LIBERO-Audio (\% success). A: GR00T; B: without binding; C: without audio modulation; D: full ListenVLA. Mean is the unweighted average over tasks; bold marks the highest mean. Each configuration was evaluated over 50 episodes per task.}
\label{tab:ablation}
\resizebox{0.7\linewidth}{!}{
\begin{tabular}{lcccc}
\toprule
Task & A & B & C & D (Full) \\
\midrule
T1 & 22.0 & 42.0 & 64.0 & 64.0 \\
T2 & 36.0 & 36.0 & 80.0 & 96.0 \\
T3 & 38.0 & 30.0 & 84.0 & 86.0 \\
T4 & 42.0 & 28.0 & 66.0 & 94.0 \\
T5 & 6.0 & 24.0 & 18.0 & 32.0 \\
T6 & 70.0 & 80.0 & 96.0 & 94.0 \\
T7 & 28.0 & 54.0 & 48.0 & 44.0 \\
T8 & 8.0 & 12.0 & 18.0 & 20.0 \\
\midrule
Mean & 31.25 & 38.25 & 59.25 & \textbf{66.25} \\
\bottomrule
\end{tabular}
}
\end{table}

\subsection{Audio Swap Test}
Task success alone does not establish whether a policy uses sound. A policy may select the correct box by chance or follow a learned reaching pattern, as illustrated by the nonzero success of GR00T without audio (Table~\ref{tab:ablation}). We used an audio swap test as a lightweight diagnostic of the effect of sound on action generation. The test replaces recorded audio while holding images, instructions, robot state, and sampling noise fixed, isolating audio-induced changes in predicted actions.

For each task, we selected the first three eligible demonstration episodes in episode-index order, each containing target-associated and alternative anchors with complete two-second audio histories. Each episode supplied two fixed observations, giving six observations per task and 48 in total, all drawn from the training demonstrations. At each observation, we compared predictions conditioned on the two recorded audio histories.

All four methods were evaluated at the same anchor locations using their native observation interfaces. Within each model, non-audio inputs and sampling noise were held fixed while the causal audio inputs were substituted.

We measured the mean absolute difference between the decoded action chunks:
\begin{equation}
\begin{aligned}
D_{\mathrm{swap}}(o)
={}&\frac{1}{7H}\Bigl\|
\hat{A}_\theta(o,W^a;\epsilon)\\
&\qquad{}-\hat{A}_\theta(o,W^b;\epsilon)
\Bigr\|_1.
\end{aligned}
\label{eq:swap}
\end{equation}
Here, $H=16$, and $o$ denotes the fixed non-audio inputs available to each model. The norm sums absolute differences over the 16-step, seven-dimensional decoded action chunk before clipping or gripper thresholding. We averaged the six observations within each task and then the eight task means with equal weight. The score in Eq.~\eqref{eq:swap} measures audio-induced action changes rather than task correctness.

\begin{table}[t]
\centering
\caption{Audio swap test. Values are eight-task macro-averages of $D_{\mathrm{swap}}$ in decoded controller-command coordinates. Each task contributes six observations from three demonstrations, with one noise seed.}
\label{tab:audio-swap}
\resizebox{0.5\linewidth}{!}{
\begin{tabular}{lc}
\toprule
Model & Audio swap \\
\midrule
GR00T & 0.00000 \\
HEAR & 0.02931 \\
ManiWAV & 0.05812 \\
ListenVLA & 0.06214 \\
\bottomrule
\end{tabular}
}
\end{table}

ListenVLA's actions changed under audio substitution, whereas the no-audio GR00T baseline was invariant (Table~\ref{tab:audio-swap}). ListenVLA showed nonzero mean changes on all eight tasks. These results show audio-dependent action generation on the sampled observations; different native histories limit cross-model comparisons of response magnitude.

\subsection{Real-World Experiments}
\begin{figure}[t]
\centering
\includegraphics[width=0.65\linewidth]{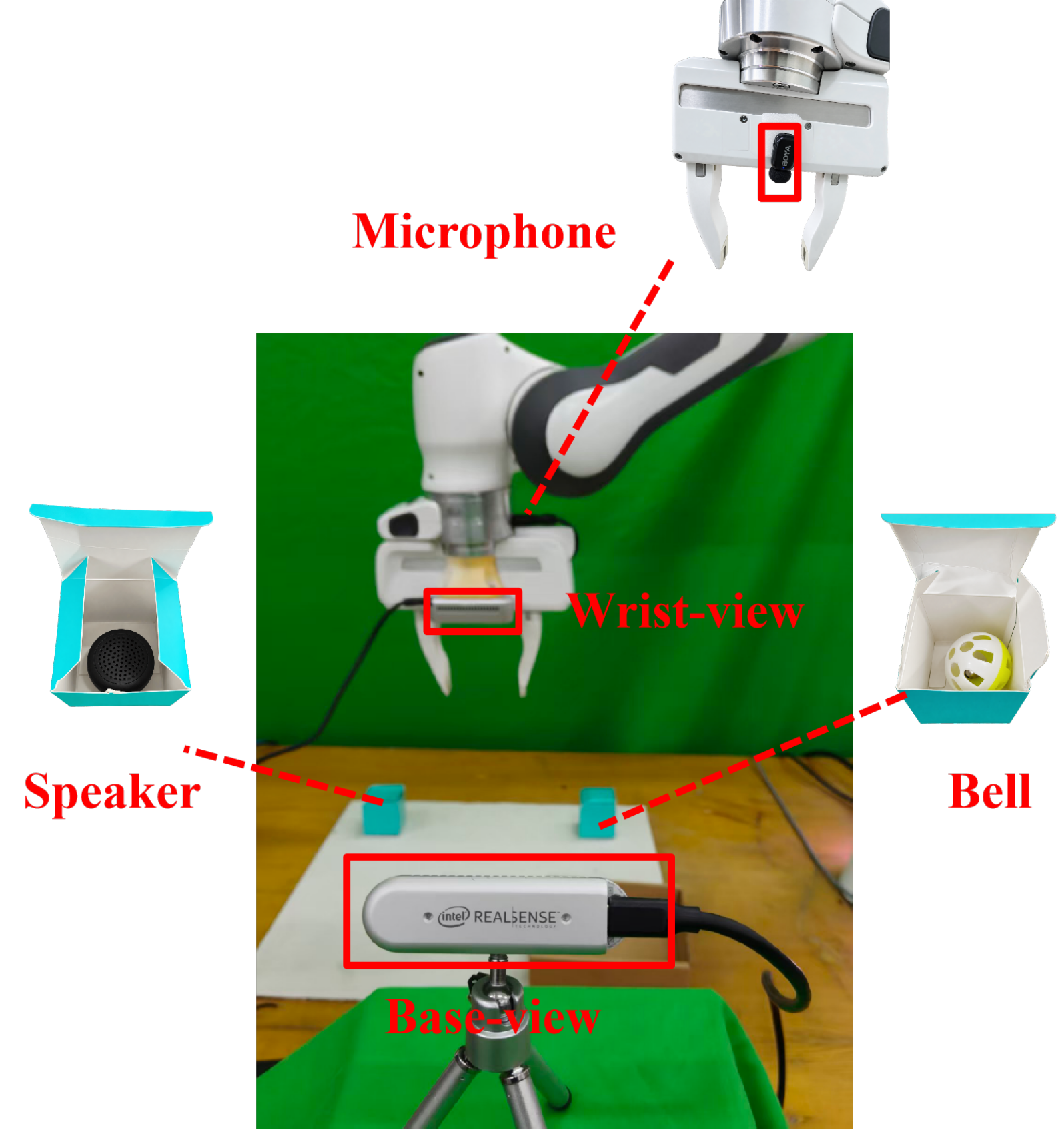}
\caption{Real-world experimental setup. The Franka Panda is equipped with external and wrist camera views and a gripper-mounted microphone. Insets show the microphone and the speaker and bell enclosed in the task objects.}
\label{fig:real-setup}
\end{figure}

We evaluated ListenVLA using a Franka Panda arm, an Intel RealSense D435 camera, and a BOYA mini microphone positioned on opposite sides of the arm (Fig.~\ref{fig:real-setup}). The setup provided external and wrist camera views, with the microphone mounted near the gripper. Models were fine-tuned with task-specific demonstrations. We performed 15 evaluation trials for each of three tasks.

\paragraph{R1: Sounding-box selection}
Two boxes were placed on the table, with one playing a beep. The robot had to identify and pick up the sounding box.

\paragraph{R2: Delayed-beep response}
The robot had to remain stationary until a box began playing a beep and then pick it up. Sound onset was randomized between 3 and 12 seconds after each trial began. Any robot motion before sound onset or failure to initiate motion within 2 s afterward was counted as a failure, even if the box was subsequently retrieved.
\paragraph{R3: Bell-containing-box identification}
Two boxes were initially placed on the table, one containing a bell. The robot had to grasp and shake the boxes to obtain auditory evidence, then place the bell-containing box in the receptacle. During deployment, we scaled the action commands used for shaking to increase the motion amplitude.

\begin{table}[t]
\centering
\caption{Real-world task completion after task-specific fine-tuning. Each task was evaluated over 15 trials.}
\label{tab:real-world}
\resizebox{\linewidth}{!}{
\begin{tabular}{lcc}
\toprule
Task & Successes / trials & Success rate (\%) \\
\midrule
R1: Sounding-box selection & 11 / 15 & 73.3 \\
R2: Delayed-beep response & 13 / 15 & 86.7 \\
R3: Bell-containing box & 9 / 15 & 60.0 \\
\bottomrule
\end{tabular}
}
\end{table}

ListenVLA completed 11 of 15 sounding-box trials, 13 of 15 delayed-beep trials, and 9 of 15 bell-identification trials (Table~\ref{tab:real-world}). These tasks tested target selection, response timing, and interaction to reveal auditory evidence.

\begin{figure*}[t]
\centering
\includegraphics[width=0.80\textwidth]{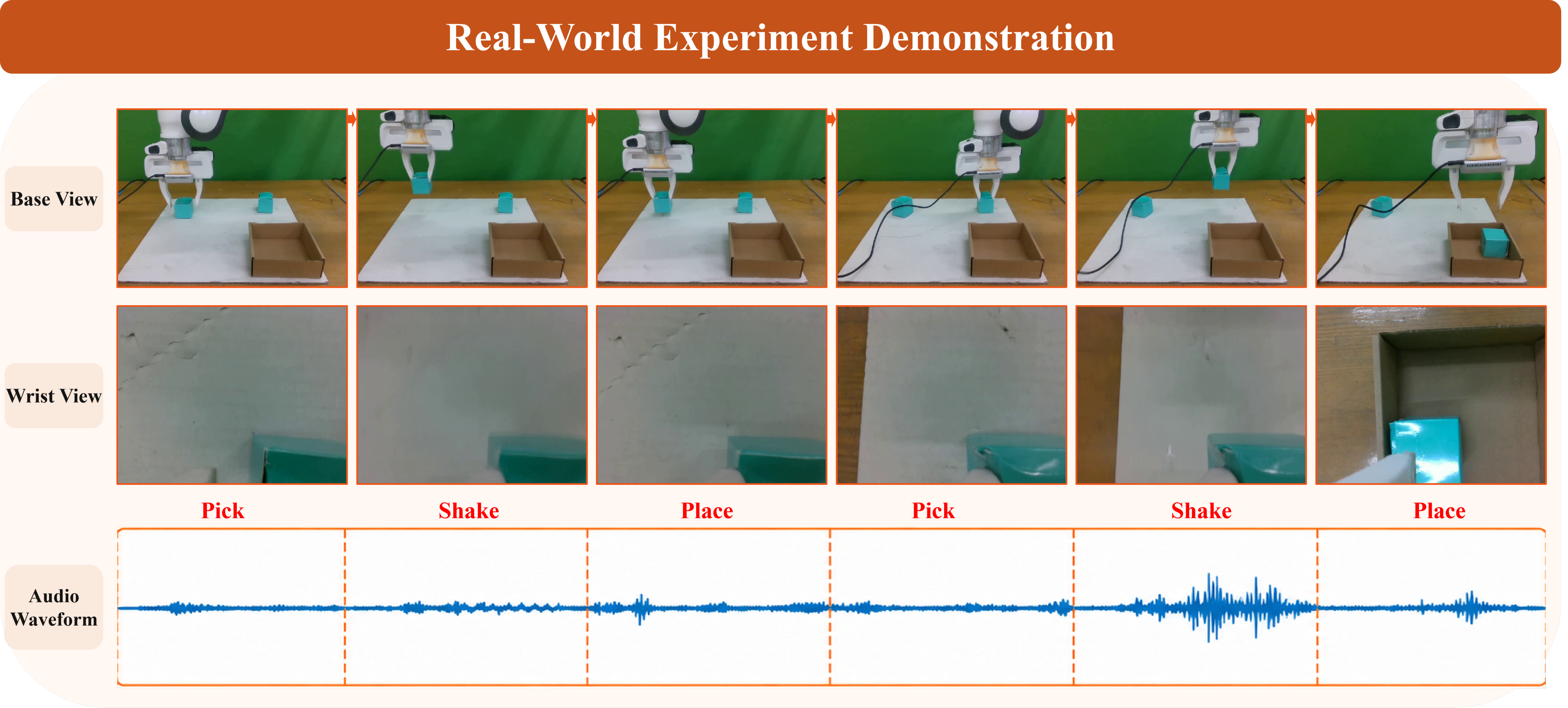}
\caption{Real-world rollout for bell-containing-box identification (R3), shown from left to right. Columns 1--3 show grasping and shaking the first box and returning it to the table after no bell sound was heard. Columns 4--6 show grasping and shaking the second box, eliciting a rattle, and placing it in the receptacle. The rows show external-camera images, wrist-camera images, and the corresponding audio waveform.}
\label{fig:real-rollout}
\end{figure*}

Fig.~\ref{fig:real-rollout} illustrates an R3 rollout in which the robot returned the first box and retrieved the second after hearing its bell. The sequence connects the probing actions with auditory feedback and subsequent target selection.

\section{Conclusion}
We presented ListenVLA, an audio-conditioned manipulation policy built on GR00T N1.7, together with LIBERO-Audio, an eight-task benchmark for sound-guided manipulation. ListenVLA combines causal audio conditioning with an audio–action binding objective that links auditory differences to demonstrated action differences. Experiments on LIBERO-Audio showed higher average task success than the evaluated audio-conditioned baselines. Ablation studies supported the contributions of binding supervision and audio modulation, while audio swap tests showed that changing audio influences action predictions. After task-specific fine-tuning, ListenVLA also performed sound-guided manipulation in three real-world tasks. Future work will investigate how robots can more effectively acquire auditory evidence through interaction and use it to guide subsequent actions.

\ifCLASSOPTIONcaptionsoff
  \newpage
\fi

\bibliographystyle{IEEEtran}
\bibliography{reference}



\end{document}